\documentclass[letterpaper,twocolumn,10pt]{article}
\usepackage{usenix2019_v3}
\usepackage{tikz}
\usepackage{amsmath}
\usepackage{listings}
\usepackage{xcolor}
\usepackage{hyperref}
\hypersetup{
    colorlinks=true,   % false: boxed links; true: colored links
    linkcolor=blue,    % color of internal links
    citecolor=green,   % color of links to bibliography
    urlcolor=magenta   % color of external links (like ORCID)
}
 
\definecolor{codegreen}{rgb}{0,0.6,0}
\definecolor{codegray}{rgb}{0.5,0.5,0.5}
\definecolor{codepurple}{rgb}{0.58,0,0.82}
\definecolor{backcolour}{rgb}{0.95,0.95,0.92}
 
\lstdefinestyle{mystyle}{
    backgroundcolor=\color{backcolour},   
    commentstyle=\color{codegreen},
    keywordstyle=\color{magenta},
    numberstyle=\tiny\color{codegray},
    stringstyle=\color{codepurple},
    basicstyle=\ttfamily\footnotesize,
    breakatwhitespace=false,         
    breaklines=true,                 
    captionpos=b,                    
    keepspaces=true,                 
    numbers=left,                    
    numbersep=5pt,                  
    showspaces=false,                
    showstringspaces=false,
    showtabs=false,                  
    tabsize=2
}
 
\begin{document}
%-------------------------------------------------------------------------------

\date{}

\title{\Large \bf KernelOracle: Predicting the Linux Scheduler's Next Move with Deep Learning}

\author{
  {\rm Sampanna Yashwant Kahu} \\
  {\rm Department of Electrical and Computer Engineering} \\
  {\rm Virginia Polytechnic Institute and State University} \\
  {\rm Blacksburg, VA 24061, USA} \\
  \texttt{sampanna@vt.edu} \\
  ORCID: \href{https://orcid.org/0000-0002-8522-2926}{0000-0002-8522-2926}
}

\maketitle

%-------------------------------------------------------------------------------
\begin{abstract}
%-------------------------------------------------------------------------------
Efficient task scheduling is paramount in the Linux kernel, where the Completely Fair Scheduler (CFS) meticulously manages CPU resources to balance high utilization with interactive responsiveness. This research pioneers the use of deep learning techniques to predict the sequence of tasks selected by CFS, aiming to evaluate the feasibility of a more generalized and potentially more adaptive task scheduler for diverse workloads. Our core contributions are twofold: first, the systematic generation and curation of a novel scheduling dataset from a running Linux kernel, capturing real-world CFS behavior; and second, the development, training, and evaluation of a Long Short-Term Memory (LSTM) network designed to accurately forecast the next task to be scheduled. This paper further discusses the practical pathways and implications of integrating such a predictive model into the kernel's scheduling framework. The findings and methodologies presented herein open avenues for data-driven advancements in kernel scheduling, with the full source code provided for reproducibility and further exploration. \\

The source code and data used in this work is available at: https://github.com/SampannaKahu/KernelOracle.
\end{abstract}

%-------------------------------------------------------------------------------
\section{Introduction (Problem statement)}
%-------------------------------------------------------------------------------

Currently, the process scheduling in the Linux kernel is done by the CFS. However, CFS does not recognise each process based on it's previous execution history. Basically, it's objective is to best utilize processor time. However, the Linux kernel does not take into account the past execution history of the processes into account when scheduling a process. It also tries to preempt a running process even it is about to finish it's execution which leads to multiple context switches. However, if a scheduler could take into account the execution history of a process and if it could also predict a process is going to complete it's execution on the processor, it could let the process finish and then schedule the next process. This could improve the overall performance of the kernel since this could lead to lesser context switches. \\

The possibilities of this work are numerous. For example, if the scheduler could predict the amount of time a process is going to execute, it could make a well-informed decision about whether to schedule a process at a certain time or not. Further, the prediction of process execution could be extended to any domain where scheduling is used. For example, in Android, we know that opening some apps take significant amount of time because the app's data and resources need to loaded into the memory. However, if the scheduler could predict this in advance, the necessary data could be pre-fetched into the memory to ensure a better user experience when using the device.

%-------------------------------------------------------------------------------
\section{Background}
%-------------------------------------------------------------------------------

This section talks about the background about the various concepts and tools used in this work.\\

Almost all machines these days are multi-processor machines. This means the multiple processors are executing tasks parallely. This brings in the notion of hardware threads. As explained in \cite{hardware_threads_wikipedia}, hardware threads are capable of handling a single process. This means that, the hardware is controlling the threading. Completely Fair Scheduler (CFS) decides which process will be executed next and the execution of which process will be preempted. As mentioned in the official documentation \cite{cfs_official_documentation}, CFS basically models an "ideal, precise multi-tasking CPU" on real hardware. On real hardware, only a single task can be run at any given instant. Thus, it is necessary to introduce the concept of "Virtual Runtime" which specifies when the next time-slice of a task would start execution on an ideal multi-tasking CPU \cite{cfs_official_documentation}.\\

The CFS is triggered every-time the system-clock/timer is triggered. A scheduling routine runs on each timer shot and checks whether any new process needs to be scheduled. This mechanism uses multiple data structures some of which are task-struct, queue of waiting processes, the rb-tree which signifies which task has spent how much amount of time on the CPU, and so on. However, this all happens internally in the kernel and no concrete data is available to get better visibility in this scheduling process. Thus, many profiling tools have been built to collect these kind of data. One of such tools is the \textit{perf-sched} \cite{brendan_gregg_perf} tool. This tool uses the internal kernel functions and collects metrics from the using kprobes and kretprobes. Each version of the \textit{perf-sched} tool is specific to a certain kernel because the function signature of the kernel change with each new version. The data returned by this tool can be stored in \textit{.csv} format and sued for further downstream analysis.\\

NGinx \cite{what_is_nginx_website} is a popular open-source software for high-performance web-servers. It is also commonly used as a proxy and sometimes as a load-balancer as well. It is one of the fastest server architectures available openly and constantly beats the performance of the Apache LAMP stack. Because of its performance and popularity it is commonly used in academic research papers as part of experiments \cite{lwcs_paper}.\\

Ab \cite{apache_bench_wikipedia}, also known as ApacheBench, is a  popular tool by Apache for simple bench-marking and load-testing of http endpoints. This is a command-line tool and can send numerous parallel requests to any given http endpoint. Due to its easy setup, it is the logical choice to use it in experiments. \\

LSTMs are a popular tool for modelling time-series data, or any general sequence of data. They were first proposed by Hochreiter and Schmidhuber in 1997 in their famous paper \cite{lstm_paper}. LSTMs use forget and remember gates for better remembering both the long and short term instances of data they encountered. Further, they are easier to train that a regular recurrent neural network because there is an easy/shorter path for gradients to flow backwards between the LSTM cells.

%-------------------------------------------------------------------------------
\section{Related work}
%-------------------------------------------------------------------------------

In 2005, Negi. et. al, proposed a new method for predicting the next process to be scheduled. They used machine learning techniques to learn the CPU time-slice utilization behaviour of known processes with the goal to minimize the TAT (turn around time) of a process. After collecting the data, they used decision trees to predict the next task to be scheduled. However, our work differs from their because we use deep learning techniques (LSTMs) to model the CPU time-slice utilization behaviour instead of a decision tree. Although LSTMs might be computationally heavier than decision trees, the aim of our work is to evaluate their performance for this particular application.\\

An interesting application of machine learning towards Linux kernel was demonstrated by Sasha Levin and Julia Lawall in 2018 \cite{ml_and_stable_kernels}. They used machine learning to identify patches that fix bugs from the ones that don't. As input to their machine learning model, they used the source code of the patch. Then, based on the past history of which patches were applied to the stable tree, they created the ground truth labels and trained their machine learning model. As a result, their model was able to classify patches with an accuracy of more than 80\%. However, this is not a direct application of machine learning inside the running Linux kernel and hence is different that our work here.\\

Some work on predicting schedulers was done by Shin et. al. in 2012 \cite{predicting_mobile_app_usage}. As a part of this work, they predicted the usage patterns of applications running on a mobile phone. Their model predicted the usage of 9 candidate apps with an accuracy of over 87\%.

%-------------------------------------------------------------------------------
\section{Motivation}
%-------------------------------------------------------------------------------

As mentioned in the previous section, the model developed in this work could be used to enhance the Linux scheduler. For example, if the model could predict the amount of time it is going to spend on the processor, the CFS scheduler could take a better informed decision about when to schedule this task. Further, if the model could predict how much more time a process is going to spend on the scheduler, it could preempt the process a little later or a little earlier depending upon the requirement. This could save a lot of time spent in context switching. This could speed up the Linux kernel considerably.\\

Further, this type of model could be potentially applied to other types of scheduling problems too. For example, in case of the Android mobile operating system, a number of applications can be installed and executed. However, it takes considerable time for some applications to load it's resources in the memory and start executing. An analogy could be drawn to the context switching between two processes which consumes vital CPU resources. A model which could predict the next task that is going to be executed could pre-load it's resources into the memory to save time and computation \cite{prodiction_next_app_android}.

%-------------------------------------------------------------------------------
\section{Methodology}
\label{sec:figs}
%-------------------------------------------------------------------------------

\begin{figure}
\begin{center}
\includegraphics[width=\columnwidth]{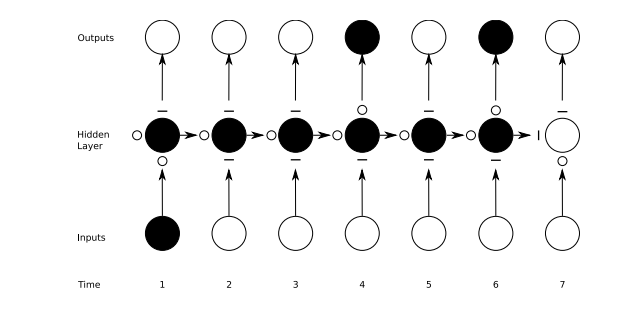}
\end{center}
\caption{\label{fig:rnn} High-level structure of Recurrent Neural Networks \cite{rnn_image}.}
\end{figure}

\subsection{Dataset generation:}

For data generation, we used the \textit{perf} tool available in Linux. Via this tool, it is possible to get detailed metrics around scheduling of the task. This tool needs to be run in root(\textit{sudo}) mode. This tool generates a \textit{perf.data} file which is a compressed version of all the collected data during the session. This file can be opened using the tool itself.\\

Following command is used to record the task scheduler metrics of the current kernel:

\begin{lstlisting}[language=Bash, caption=Command to record the scheduling data from the current running kernel for a total of 50 seconds. The output will be stored in a file in the same folder with the name \textit{perf.data}.]
sudo perf sched record -- sleep 50
\end{lstlisting}

\begin{figure}
\begin{center}
\includegraphics[width=\columnwidth]{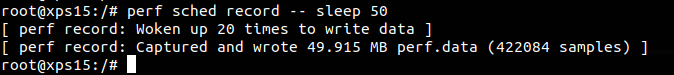}
\end{center}
\caption{\label{fig:perf_sched_record} Example of recording scheduler metrics using the \textit{perf} command. This example test was run for 50 seconds. A total of 49.915 MB of data was collected which consisted of 422084 samples.}
\end{figure}

This command shown in Figure \ref{fig:perf_sched_record} will record the scheduling metrics for all the cores and all the processes in the current running kernel. As shown in the figure, a total of about 50 MB of data was collected in 50 seconds which consisted of about 422084 samples. However, it is worth noting that the amount of data collected depends the load of the system and the number of processes being scheduled. For example, if a lot of processes are getting scheduled on the system, the number of samples in the collected profiler data using \textit{perf} will be higher.

\begin{figure}
\begin{center}
\includegraphics[width=\columnwidth]{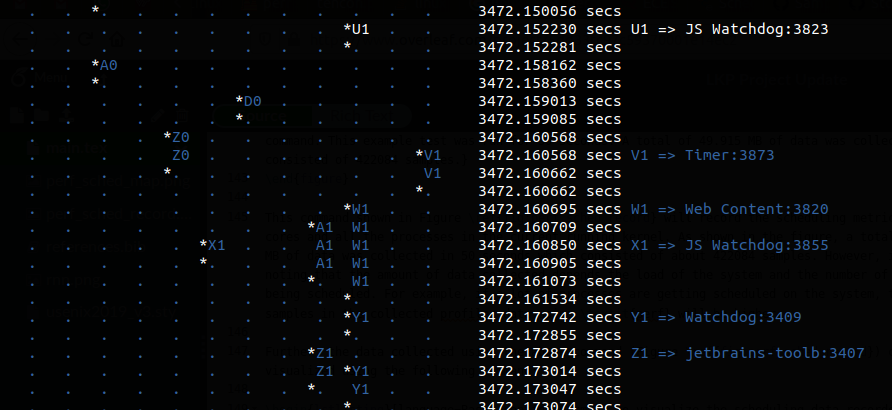}
\end{center}
\caption{\label{fig:perf_sched_map} Visualization of the 'perf sched' command.}
\end{figure}

Further, the data collected using the above command (Figure \ref{fig:perf_sched_record}) can be visualized using the following command:

\begin{lstlisting}[language=Bash, caption=Command to visualize the scheduling data recorded using the record command. This command expects a file named \textit{perf.data} in the current directory.]
sudo perf sched map
\end{lstlisting}

The output of the above command can be seen in Figure \ref{fig:perf_sched_map}.

%-------------------------------------------------------------------------------
\subsection{Experimental setup}
%-------------------------------------------------------------------------------

\subsubsection{Hardware parameters and operating system}
As evident in Figure \ref{fig:perf_sched_map}, the visualized data is for a multi-CPU machine. That machine can support 12 threads in total. Each CPU is executing a single process in parallel with the other 11 CPUs. However, in this work, predicting the scheduling performance of multiple CPUs will complicate things. Thus, for this work, we will limit the scope to only a single CPU machine.\\

To accommodate this constraint, we run our experiments on a single CPU machine (i.e. a machine which supports only a single hard-ware thread). We accomplish this using a virtual machine which is configured to use only a single CPU (i.e. a single hard-ware thread). We use Ubuntu 18.04 Bionic as the operating system running in the virtual machine. This operating system is running the Linux Kernel 4.15.0-58-generic. The host operating system is Ubuntu 18.04 Bionic as well which is running the Linux Kernel 5.0.0-36-generic. The host hardware is a Dell XPS 15 laptop with a total of 12 processors of type Intel(R) Core(TM) i7-8750H CPU @ 2.20GHz.

\subsubsection{Load generation}
We tried to record the data on this virtual machine using the command shown in Figure \ref{fig:perf_sched_record}. However, the number of samples in this recording were quite low. Also, we observed that the CPU was idle for considerable amount of time. Upon further investigation, we found out that this was happening because the kernel was not under much load. In other words, not many applications were running on the operating system. This was causing the processor to sit idle intermittently. Thus, to mitigate this issue, we induced artificial load on the kernel.\\

Taking inspiration from one of the experiments in \cite{lwcs_paper}, we started the \textit{nginx} server with the default settings. However, simply starting the nginx server would not induce load on kernel. However, we tackled this by keeping the nginx server busy. Since nginx is an http web server, it will become busy if it is serving requests. Thus, it will be sufficient to keep sending requests to the nginx server to keep it busy.\\

Sending http requests to nginx was achieved by a tool called \textit{ab}. This is the acronym for Apache Bench. This tool comes as a part of the the apache2-utils package. It sends dummy http requests to a given IP address. It stops sending requests when it has finished sending the configured number of requests. We configure this tool to send http requests to nginx which is running on the same virtual machine. As a result, both nginx and ab will generate load for the processor. We start this tool using the following command:

\begin{lstlisting}[language=Bash, caption=Command to send one million requests to the given IP address using ab (Apache Bench).]
ab -n 1000000 -c 5 http://10.0.2.15/
\end{lstlisting}

With the given configuration of hardware (single processor i7 virtual machine), it takes more than 100 seconds for these 1 million http requests to complete. Thus, we record the scheduler profiler data using the command mentioned in Figure \ref{fig:perf_sched_record} during this time-frame.

\subsubsection{Data pre-processing}
\begin{figure}
\begin{center}
\includegraphics[width=\columnwidth]{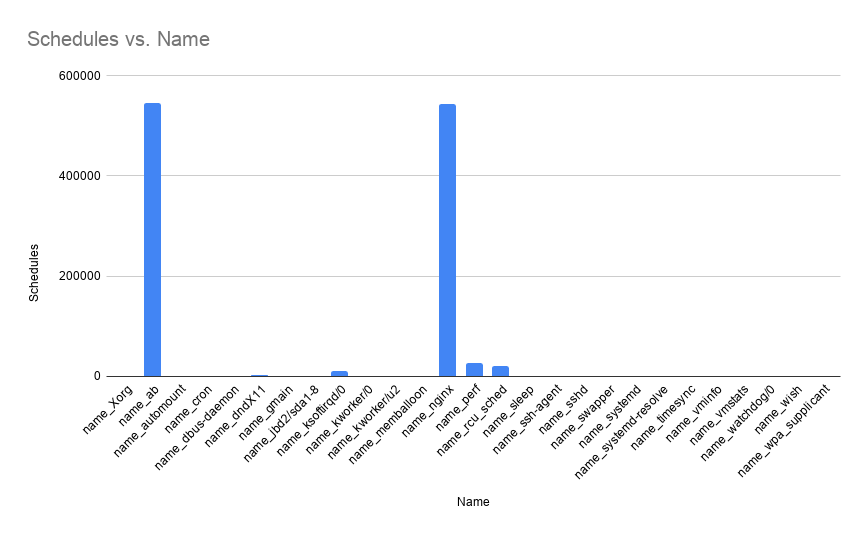}
\end{center}
\caption{\label{fig:schedules_vs_name} The number of schedules of a given process. This was captured when 'ab' was generating load on 'nginx' by sending a total of 1 million http requests.}
\end{figure}

The data collected by the the \textit{perf} tool is not suitable to be given as an input to the machine learning model. In it's raw format, it encodes the task names into alpha-numeric format based on occurrence (see Figure \ref{fig:perf_sched_map}). Thus, we need to convert these alphanumeric code into their real task names. To do so, we write a simple python script to read the mapping between each alphanumeric code with it's corresponding task name and replace those codes with their real task names. The initial statistics of the collected dataset can be seen in Figure \ref{fig:schedules_vs_name}.\\

Further, these task names are categorical data. They are currently stored in string format because they are names. We convert them into vectors by applying one-hot encoding on these names. In the dataset that we collected, we found a total of 28 task names during the entire recording process. Thus, the dimension-size of these one-hot vectors would be 28 after applying the one-hot encoding.\\

The data recorded by \textit{perf} also contains the time-stamp at which a process was scheduled. The significant/meaning of the absolute value of this time-stamp is not clear. However, all we are interested for the scope of this work is the time duration between two consecutive schedules of processes. This data can be easily obtained from the time-stamps by differentiating them. In essence, we take the differences of consecutive time-stamps to differentiate them. We drop the last record. Further, we scale and shift this data to get zero mean and unit variance.\\

One unexpected issue encountered during the pre-processing of the data was that one of the differentiated time values was abnormally higher than the others. Its value was more than one complete second. This meant that the corresponding process took more than 1 second to be scheduled, execute its time-slice and get preempted. However, that time duration is unusually high for any process in the kernel. Hence, we got rid of that entry from the dataset assuming that it was a one-off issue.

\subsection{Model:}
\begin{figure}
\begin{center}
\includegraphics[width=\columnwidth]{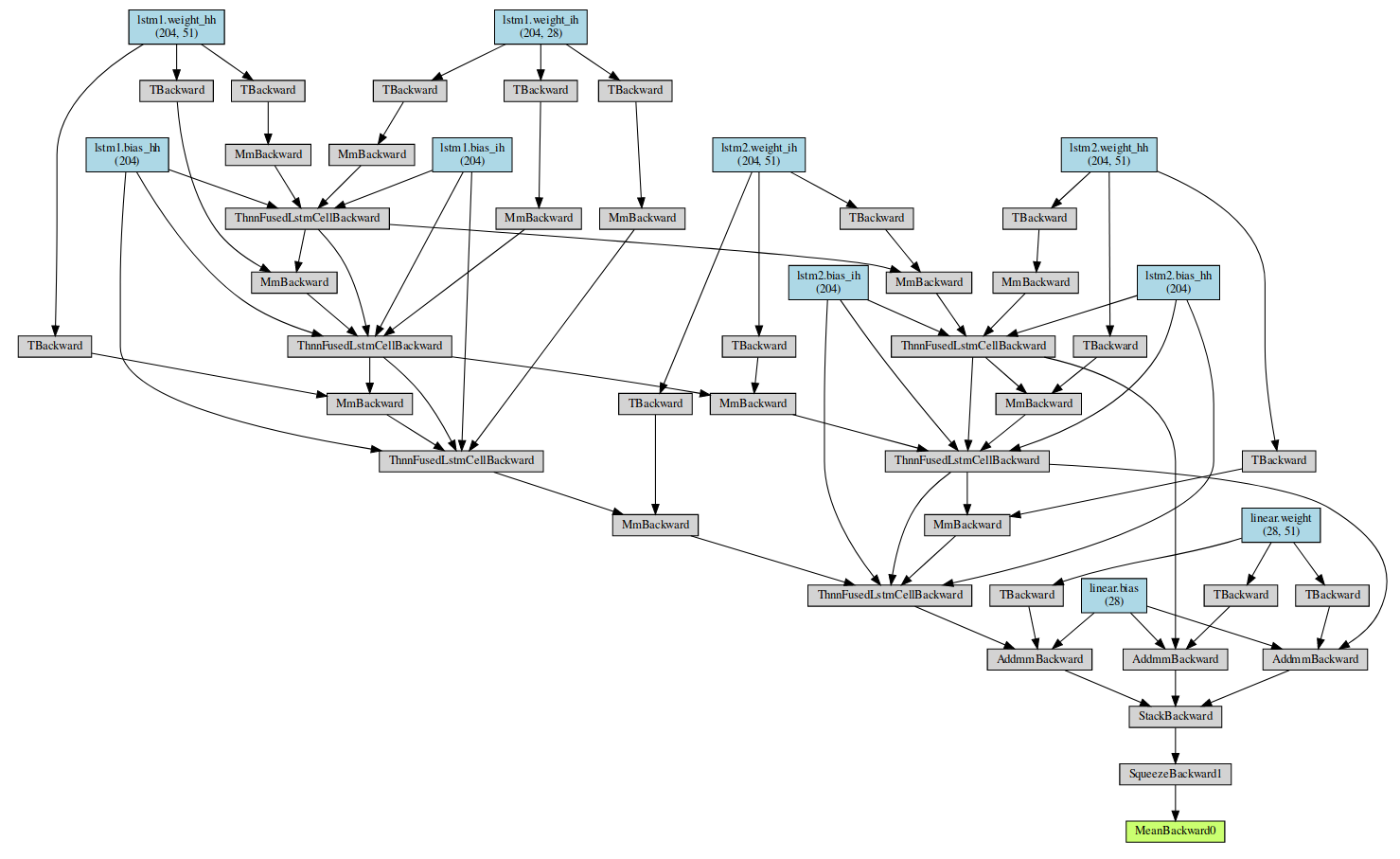}
\end{center}
\caption{\label{fig:model_graph} The computational graph of our model.}
\end{figure}

The collected dataset is a series of readings collected over time. Which means that this is a time-series dataset. According to the latest research in the field, Deep Learning models are the best fit of handling large amounts of data. Specifically, the Long-Short-Term memory (LSTMs) \cite{lstm_original_paper, lstm_paper} (a type of Recurrent Neural Networks (RNNs) \cite{rnn_paper} See figure \ref{fig:rnn}) are best suited to handle time-series data.\\
We train an LSTM model using this data to predict the next task that is going to be scheduled or preempted. It is important to note that the model is trained on the difference of the consecutive time intervals between two schedules. In other words, we record the time intervals between two consecutive schedules and differentiate them. Further, we select sequences of appropriate length for training the model. The model is trained until the test loss reasonably stabilizes. Further, we use the LGFBS optimizer for 30 epochs or until the loss converges. See figure \ref{fig:model_graph} for the computational graph of our LSTM model generated using PyTorchViz \cite{pytorchviz_github}.\\

%-------------------------------------------------------------------------------
\section{Results}
%-------------------------------------------------------------------------------

\begin{figure}
\begin{center}
\includegraphics[width=\columnwidth]{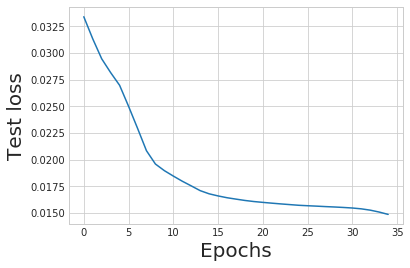}
\end{center}
\caption{\label{fig:test_loss} Plot of test loss vs. the epochs.}
\end{figure}

We computed and recorded the test loss for each epoch during the training process. Figure \ref{fig:test_loss} shows the test loss as the training process progresses. From this figure, we can observe that initially the test loss was about 0.0325. However, as the training progresses, the loss started to reduce and stabilized after 30 epochs. Beyond this point, training further won't lead to the further decrease in the loss and would simply lead to over-fitting. For our evaluations, we use the model saved at 30 epochs because it has the least test loss.\\

After the training process was complete, we provide sample input sequences to the trained model. We make the model generate predictions after the sample input sequence is over. In other words, once the model sees the entire sample input sequence, we run the model further for 1000 steps to make it predict the next schedules. Also, to get a better idea of how the model learns to predict the sequences, we make the model generate predictions after each epoch.\\

\begin{figure}
\begin{center}
\includegraphics[width=\columnwidth]{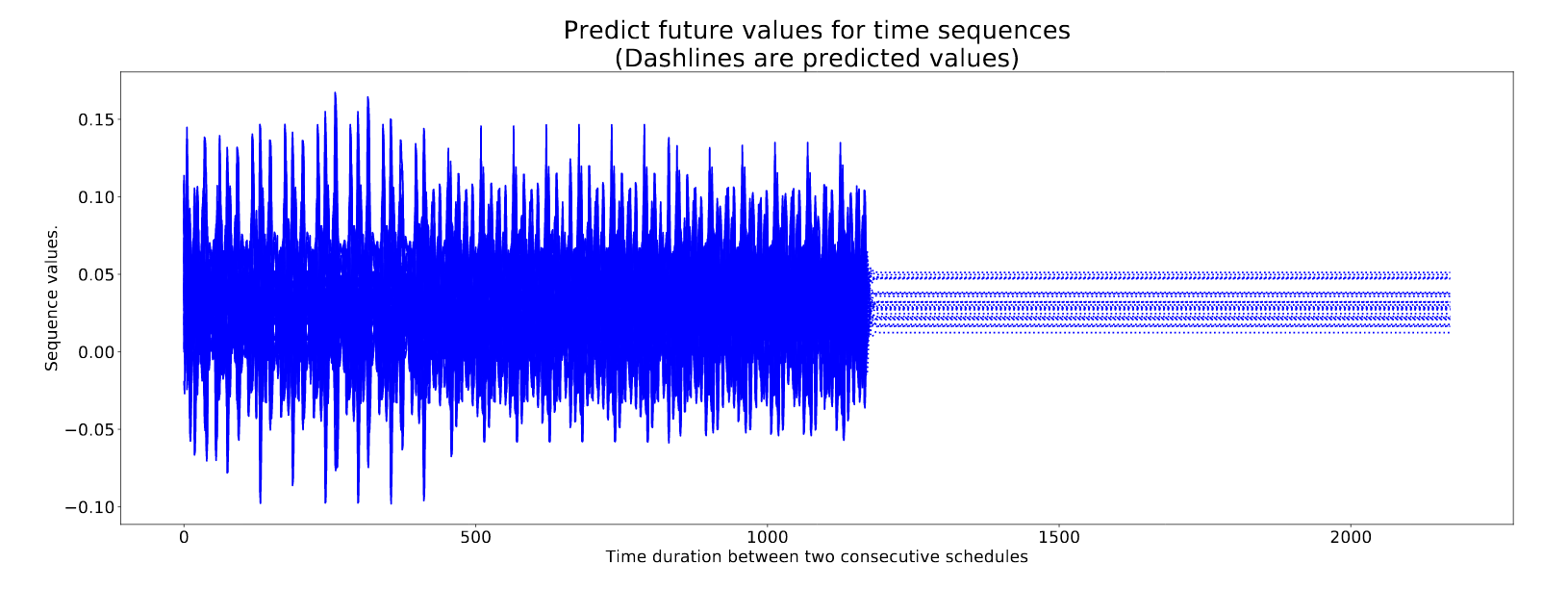}
\end{center}
\caption{\label{fig:result_0} Prediction of the model after training the model for 1 epoch.}
\end{figure}

Figure \ref{fig:result_0} shows the prediction of the model for a sample input sequence after the model was trained for 1 epoch. The values in this figure before the 1200 mark on the x-axis represent the sample input sequence. The value after that represent the values predicted by the model. It is important to note that the model was trained on the difference of the consecutive time intervals between two schedules. In other words, we recorded the time intervals between two consecutive schedules and differentiated them. We can see that the model does not predict anything in figure \ref{fig:result_0}. It is just a flat line.\\

\begin{figure}
\begin{center}
\includegraphics[width=\columnwidth]{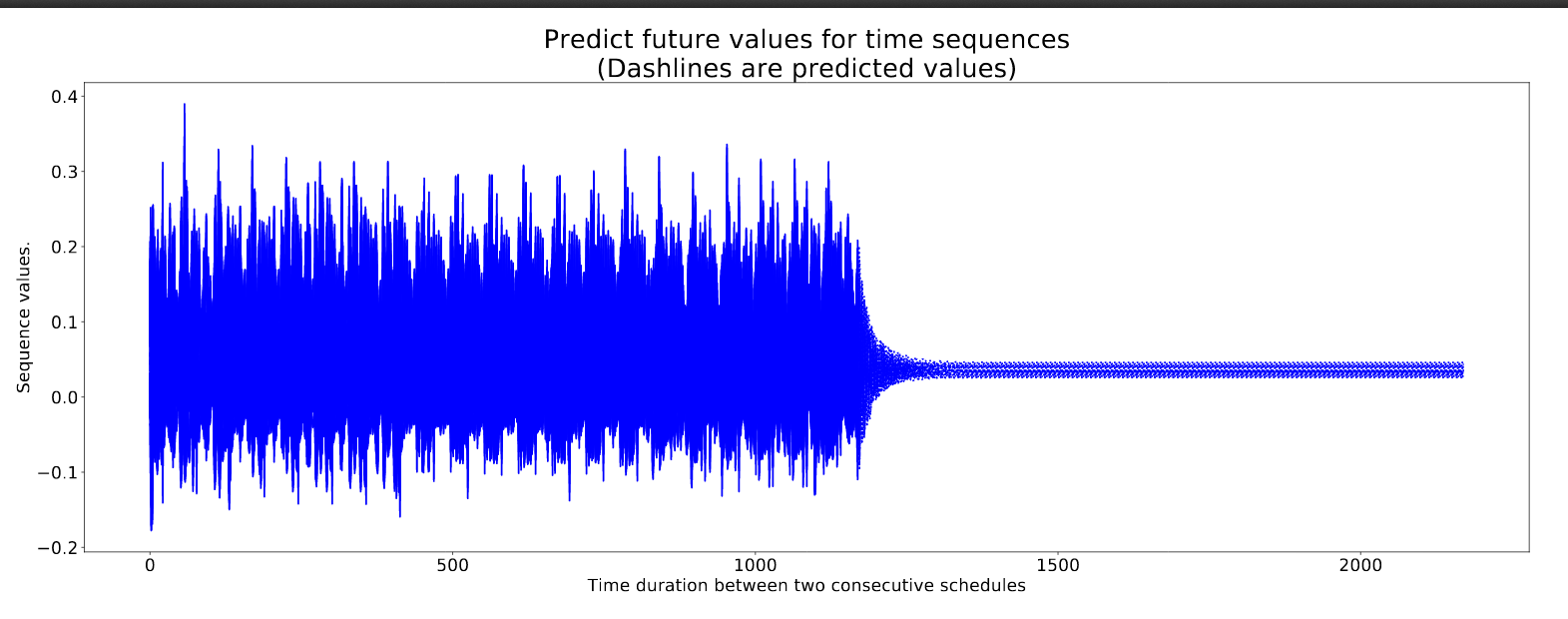}
\end{center}
\caption{\label{fig:result_2} Prediction of the model after training the model for 3 epochs.}
\end{figure}

Figure \ref{fig:result_2} shows the predictions of the model for the same sample input after 3 epochs of training. W can observe that the model is trying to match the magnitude of the training sequences.\\

\begin{figure}
\begin{center}
\includegraphics[width=\columnwidth]{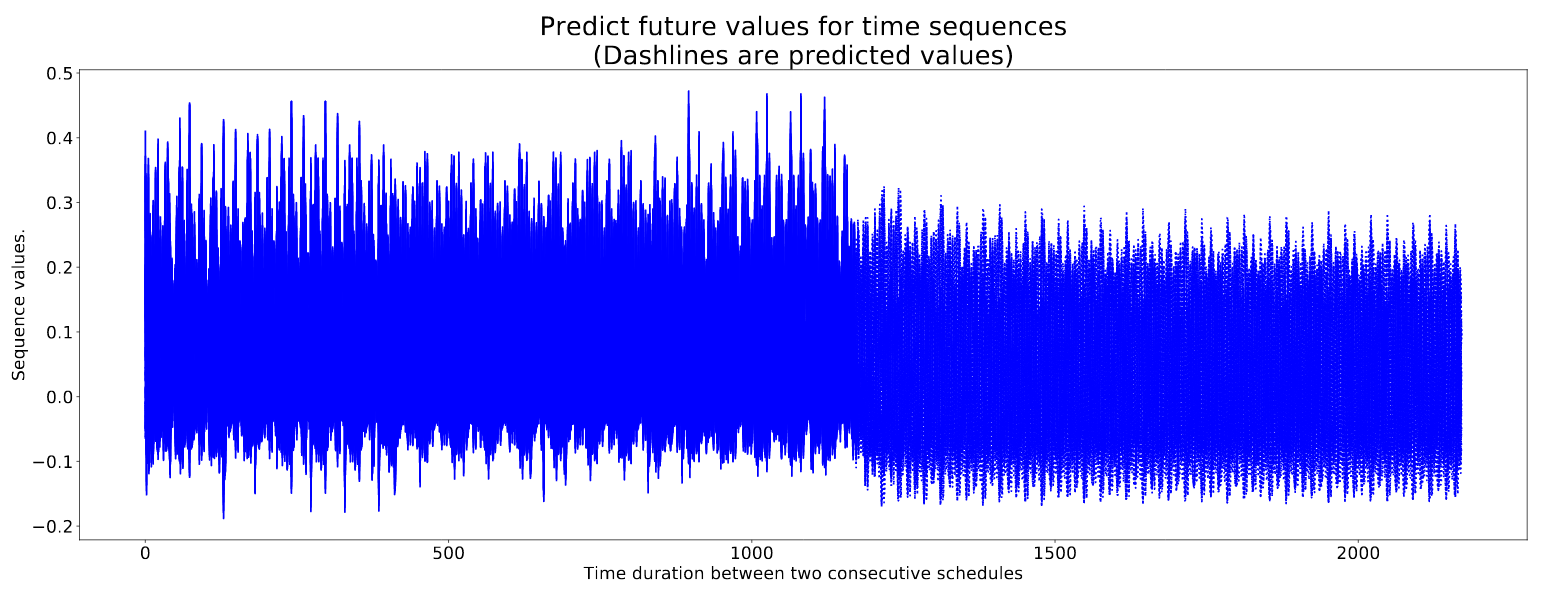}
\end{center}
\caption{\label{fig:result_4} Prediction of the model after training the model for 5 epochs.}
\end{figure}

\begin{figure}
\begin{center}
\includegraphics[width=\columnwidth]{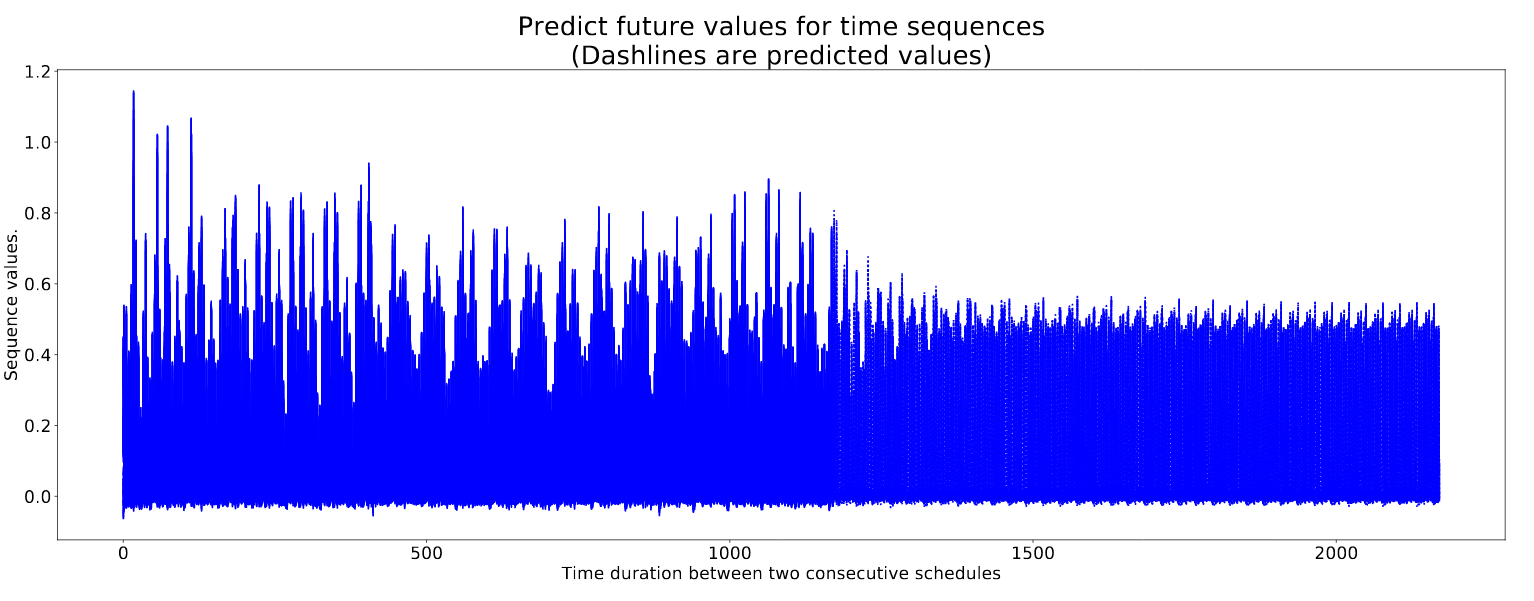}
\end{center}
\caption{\label{fig:result_14} Prediction of the model after training the model for 15 epochs.}
\end{figure}

Further, in figure \ref{fig:result_4}, the model is trained for 5 epochs. The predictions of the magnitudes is good enough. However, we can see that the model has some bias in the predictions. After a few more epochs, the model looses that bias and now the predictions look much better in figure \ref{fig:result_14}.

\begin{figure}
\begin{center}
\includegraphics[width=\columnwidth]{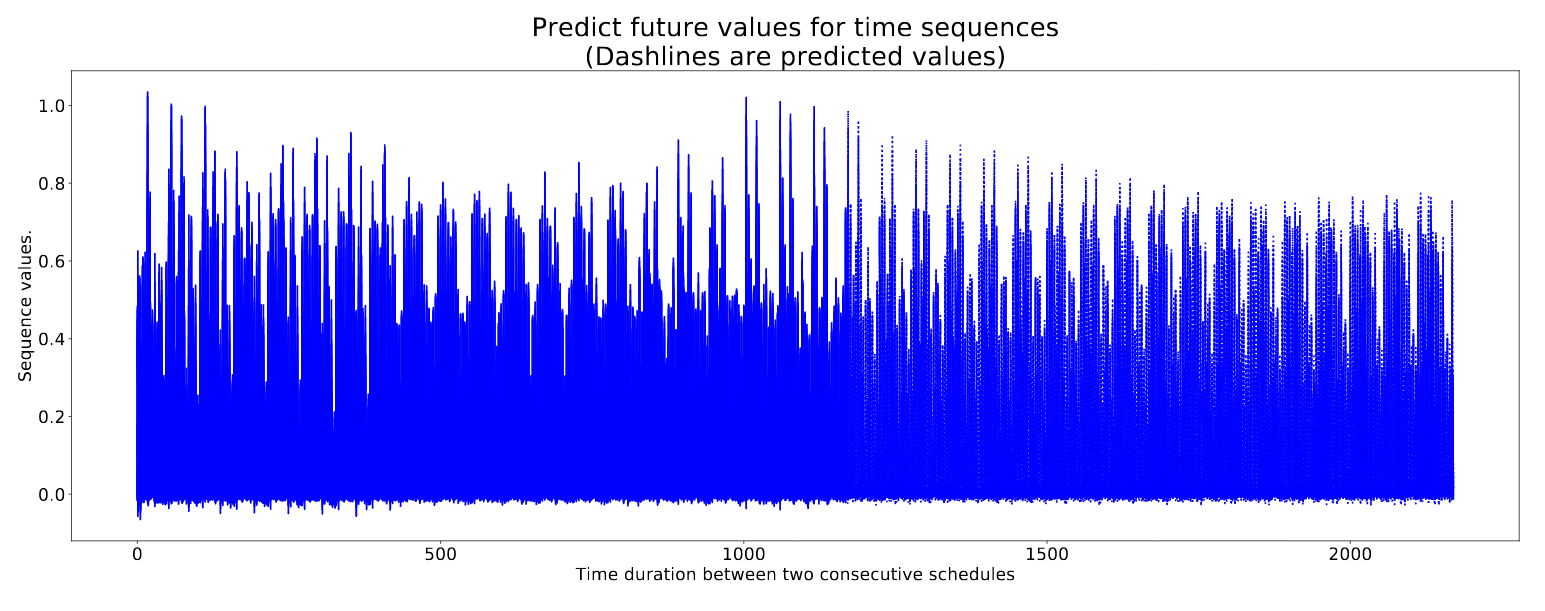}
\end{center}
\caption{\label{fig:result_29} Prediction of the model after training the model for 30 epochs.}
\end{figure}

Finally, in figure \ref{fig:result_29}, the model is now trained for 30 epochs. The magnitude of the predictions match very closely with the magnitudes of the input sequence. Further, we can observe that even the patterns of values in the predicted sequence resembles the pattern in the input sequence. This indicates that model has now been trained well which is also confirmed by stabilized test loss in figure \ref{fig:test_loss}.

%-------------------------------------------------------------------------------
\section{Discussion}
%-------------------------------------------------------------------------------

From figures \ref{fig:result_0}, \ref{fig:result_2}, \ref{fig:result_4}, \ref{fig:result_14}, \ref{fig:result_29}, we can see the progression of the predictions of the model as it is trained. Initially, the model was not predicting anything at all. It was simply using predicting a flat line. Further, as the training progressed, we can observe that the model slowly starts to acquire characteristics of the training data. For example, first, it starts to model the magnitude of the input sequences. Then, it starts to model the patterns in the input sequences. After a certain amount of training, it becomes difficult for even a human to distinguish between the sequences predicted by the model and the input sequence.\\

This capability of our model to predict the future of the scheduled tasks makes it valuable. At any given time, of the Linux kernel knows the next N tasks to be scheduled, it can take better decisions while scheduling them. For example, it could be possible to club two relatively small CPU time slices allotted for a given process into a a bigger slice. This would help reduce the number of context switches between tasks thereby improving performance.\\

To practically integrate our work into the Linux kernel, the biggest challenge is to deal with the predictions latencies of the model. Deep learning models are orders of magnitude slow when compared to the latencies of the scheduler. Since the scheduler runs at each context switch, it becomes crucial to bring down the latencies of the deep learning model if it ever has to replace the traditional scheduler.\\

Many ways have been explored till now to improve the predictions latencies of deep learning models. One of the prominent solutions is to use a better language, like C++. This approach is often used in autonomous cars where prediction latencies are important too. Another approach to dedicated hardware for running the predictions of the model (e.g. GPU, TPU, ASIC, etc). GPUs already provide a huge improvement over pure-cpu prediction latencies. Further, Google's TPUs \cite{tpu} are more specialized for deep learning. Finally, ASICs can theoretically bring down the prediction latencies by orders of magnitude and seem to be an interesting avenue to explore \cite{easic}.

\section{Development status}
%-------------------------------------------------------------------------------

\subsubsection{Completed tasks}
\begin{itemize}
    \item Literature survey, background and related work is complete.
    \item The experimental setup for data collection and generation have been figured out.
    \item The actual data collection and it's processing scripts are complete.
    \item Building, implementing and training the LSTM model on the collected dataset.
    \item Exploring/discussion the possibilities of integrating this the Linux kernel along with its implications.
\end{itemize}{}

%-------------------------------------------------------------------------------
\section{Conclusion}
%-------------------------------------------------------------------------------
In this work, we introduced the problem of predicting the scheduling of processes in the Linux kernel. Further, we discussed our motivation to develop a solution to predict the scheduling behaviour of Linux kernel and the benefits it will provide. We then discussed in detail our methodology to implement our proposed solution. This includes the process of generating the data, the experimental setup, and the details of training the deep learning model. Finally, we show our results, discuss them in detail and conclude.

%-------------------------------------------------------------------------------
\section{Future work}
%-------------------------------------------------------------------------------

To improve this further, we could use a more versatile dataset. In other words, we used only Apache Bench and NGinx to generate the load on the kernel. However, in the future, we could use a more versatile and natural data for training (for example, the scheduling data of a public server).

\bibliographystyle{plain}
\bibliography{references}

\end{document}